\documentclass[11pt]{article}
\usepackage[preprint]{acl}

\usepackage{times}
\usepackage{latexsym}
\usepackage[T1]{fontenc}
\usepackage[utf8]{inputenc}
\usepackage{microtype}
\usepackage{inconsolata}
\usepackage{amsthm}
\usepackage{amsmath}
\usepackage{amssymb}
\usepackage{graphicx}
\usepackage{array}
\usepackage{booktabs}
\usepackage{multirow}
\usepackage{colortbl}
\usepackage{xcolor}
\usepackage{url}
\usepackage{tcolorbox}
\usepackage{enumitem}
\usepackage{placeins}
\usepackage{float}
\tcbuselibrary{skins,breakable}

\newcolumntype{R}[1]{>{\raggedleft\arraybackslash}m{#1}}

\newcommand{\model}{Cast-R1}
\newcounter{idx}

\theoremstyle{plain}

\theoremstyle{definition}

\theoremstyle{remark}

\title{Learning Agentic Time Series Forecasting via Trajectory Distillation and Memory-Based Reinforcement Learning}

\author{\textbf{Xiaoyu Tao\textsuperscript{1}, Mingyue Cheng\textsuperscript{1},
  Yaguo Liu\textsuperscript{1}, Xingpeng Gao\textsuperscript{1},}\\
  \textbf{Qingsong Wen\textsuperscript{2}, Qi Liu\textsuperscript{1},
  Shijin Wang\textsuperscript{3}, Enhong Chen\textsuperscript{1}}\\[3pt]
  {\small\textsuperscript{1}State Key Laboratory of Cognitive Intelligence,
    University of Science and Technology of China}\\
  {\small\textsuperscript{2}Squirrel Ai Learning
    \quad\textsuperscript{3}iFLYTEK Research, Hefei, China}\\[3pt]
  {\small\texttt{\{txytiny,ygliu,gaoxingpeng\}@mail.ustc.edu.cn}}\\
  {\small\texttt{\{mycheng,qiliuql,cheneh\}@ustc.edu.cn}}\\
  {\small\texttt{qingsongedu@gmail.com, sjwang3@iflytek.com}}}

\hypersetup{
  pdftitle={Learning Agentic Time Series Forecasting via Trajectory Distillation and Memory-Based Reinforcement Learning},
  pdfauthor={Xiaoyu Tao, Mingyue Cheng, Yaguo Liu, Xingpeng Gao, Qingsong Wen, Qi Liu, Shijin Wang, Enhong Chen}
}

\begin{document}
\maketitle

\begin{abstract}
Recent agentic time series forecasting (TSF) approaches use large language models (LLMs) to autonomously orchestrate analytical and forecasting tools and reason about temporal patterns using the resulting evidence.
However, many such approaches rely on powerful LLMs in training-free, prompt-driven workflows, leaving open whether smaller models can learn these interaction and reasoning capabilities.
In this paper, we introduce Cast-R1, an 8B-parameter agentic forecasting model trained through trajectory distillation and multi-turn reinforcement learning with structured memory.
First, we develop a forecasting agent harness with specialized analysis and prediction tools to generate multi-turn interaction trajectories using a teacher LLM, which are subsequently evaluated and selected for supervised fine-tuning.
Second, we apply group relative policy optimization (GRPO) across multi-turn forecasting episodes, with structured memory preserving diagnostic evidence and intermediate forecasts for state construction. Rewards based on prediction accuracy and temporal structure guide tool selection, model routing, and forecast refinement. A difficulty-aware curriculum combines sample filtering and progressive sample ordering during training. Extensive experiments on multiple datasets demonstrate the effectiveness of Cast-R1. These findings highlight the potential of forecasting-specific post-training to equip smaller LLMs with multi-turn, tool-augmented forecasting capabilities within structured workflows.
\footnote{Our code is available at \url{https://github.com/Xiaoyu-Tao/Cast-R1-TS}.}

\end{abstract}

\section{Introduction}

Agentic time series forecasting (TSF) uses large language models (LLMs) to orchestrate analytical and forecasting tools and reason about temporal patterns from the resulting evidence~\cite{zhang2025alphacast,zhao2025timeseriesscientist}.
Through these interactions, an agent can characterize temporal patterns, select suitable forecasters, and refine predictions using accumulated evidence.

A prominent line of agentic forecasting research relies on powerful LLMs whose pretrained capabilities are elicited through prompting, without forecasting-specific policy training~\cite{zhang2025alphacast,zhao2025timeseriesscientist}.
Post-training has also advanced numerical forecasting and temporal reasoning~\cite{niulangtime,luo2025time}, as well as tool-augmented time series question answering~\cite{wu2026timeart}.
Building on these studies, we learn a single policy that jointly coordinates diagnostic tool selection, model routing, and numerical forecast revision using episode-local evidence.

We therefore ask: \textbf{How can forecasting-specific post-training enable a smaller LLM to coordinate tools and refine predictions using accumulated evidence?}
The first challenge, learning from demonstrations, is that forecasting benchmarks provide numerical targets rather than demonstrations of diagnostic tool use, model routing, and forecast revision.
This motivates generating, evaluating, and selecting teacher trajectories to supervise these decisions~\cite{chen2023fireact,cui-etal-2026-augur}.
The second challenge, learning from outcomes, arises because tool interactions change the evidence available to subsequent decisions, while forecasting quality is assessed only after the final prediction.
Learning from these outcomes requires preserving decision-relevant evidence across turns~\cite{hu2025hiagent} and using terminal forecasting feedback to improve the preceding decisions~\cite{wang2025ragen,cheng2025agent}.

We propose \textbf{Cast-R1}, an 8B-parameter agentic TSF model with a two-stage post-training framework.
First, we develop a forecasting agent harness with specialized time series analysis and prediction tools, enabling a teacher LLM to generate multi-turn trajectories.
We evaluate and select these trajectories for trajectory distillation through supervised fine-tuning (SFT), initializing the student's forecasting interaction policy.
The state--response demonstrations teach the student to interpret tool feedback and follow the workflow.

Second, we combine structured memory-based state construction with multi-turn GRPO to optimize the initialized policy.
The environment retains diagnostic summaries and intermediate model forecasts within each episode and incorporates them into subsequent decision states.
Within the preparation--forecasting--refinement protocol, the learned decisions are tool selection, model routing, and numerical forecast revision.
Terminal rewards for accuracy and temporal structure guide trajectory-level policy optimization across turns.
A difficulty-aware curriculum supports training.
Experimental results across diverse real-world datasets establish the effectiveness of Cast-R1, while ablation studies confirm the contributions of both SFT and RL to its performance. Together, these findings demonstrate that forecasting-specific post-training enables smaller LLMs to learn effective multi-turn decision-making and tool-use strategies for structured forecasting tasks.

Our main contributions are:
\begin{itemize}
    \item We introduce Cast-R1, an 8B-parameter agentic TSF model that learns evidence-guided, multi-turn tool-use policies through post-training within a structured workflow.
    \item We develop an agent harness to generate, evaluate, and select teacher trajectories for supervised distillation of forecasting policies.
    \item We combine forecasting-specific memory-based state construction with multi-turn GRPO to optimize tool selection, model routing, and forecast refinement using diagnostic evidence, intermediate forecasts, and accuracy and temporal-structure rewards.
\end{itemize}

\section{Related Work}

\subsection{LLM-Based Time Series Forecasting}
Surveys categorize LLM forecasting by architecture, training, and evaluation~\cite{xu2026forecastingagents}, and time-series agents by tasks, tools, and memory~\cite{chen2026timeseriesagents}.
Pretrained models learn temporal patterns at scale, providing reusable representations across domains and prediction horizons. Time-MoE~\cite{shitime} uses a mixture-of-experts architecture, while Chronos~\cite{ansari2025chronos} supports diverse forecasting settings.
LLM-enhanced methods incorporate text and context~\cite{xue2023promptcast,gruver2023large,liu2024lstprompt}. Time-LLM~\cite{jintime} reprograms time-series inputs for LLMs; TimeCMA~\cite{liu2025timecma} aligns temporal and textual modalities; TimeRAG~\cite{yang2025timerag} retrieves historical sequences.
Agentic frameworks introduce multi-step reasoning and tool interaction to coordinate forecasting decisions under evolving evidence~\cite{jalori-etal-2025-flairr,kim-etal-2026-ctrl}. TimeSeriesScientist~\cite{zhao2025timeseriesscientist} coordinates diagnosis, model selection, validation, and reporting within a unified forecasting workflow. These methods motivate adaptive forecasting through state-aware reasoning and tool interaction~\cite{shen2026metacaster}.
Beyond numerical accuracy, TemporalBench~\cite{weng2026temporalbench} evaluates historical understanding, contextual reasoning, and event-conditioned prediction.
\subsection{Learning Tool-Using Agents}
ReAct~\cite{yao2022react} interleaves reasoning, actions, and observations; FireAct~\cite{chen2023fireact} fine-tunes agents on teacher-generated trajectories. RAGEN~\cite{wang2025ragen} studies multi-turn RL for sequential decision-making in interactive environments; Agent-R1~\cite{cheng2025agent} supports step-level trajectories and context management across extended agent interactions. Augur~\cite{cui-etal-2026-augur} distills teacher-generated causal explanations for discrete trend prediction. Time-R1~\cite{luo2025time} combines supervised warm-up with RL for forecast-oriented reasoning; LangTime~\cite{niulangtime} uses PPO for autoregressive forecasting. TimeART~\cite{wu2026timeart} learns tool use from expert demonstrations for time-series question answering. ZARA~\cite{li-etal-2026-zara} grounds training-free motion time-series reasoning in knowledge and retrieval for activity recognition. HiAgent~\cite{hu2025hiagent} maintains working memory via subgoal and observation summaries. MemCast~\cite{tao2026memcast} conditions frozen LLMs on hierarchical training-set forecasting memories. Cast-R1 combines trajectory distillation with multi-turn GRPO to learn diagnostic tool selection, model routing, and numerical revision. Its forecasting environment organizes episode-local diagnostics and intermediate forecasts as structured memory, while accuracy and temporal-structure rewards guide policy optimization within a preparation--forecasting--refinement protocol.

\begin{figure*}[t]
    \centering
    \includegraphics[width=1\linewidth]{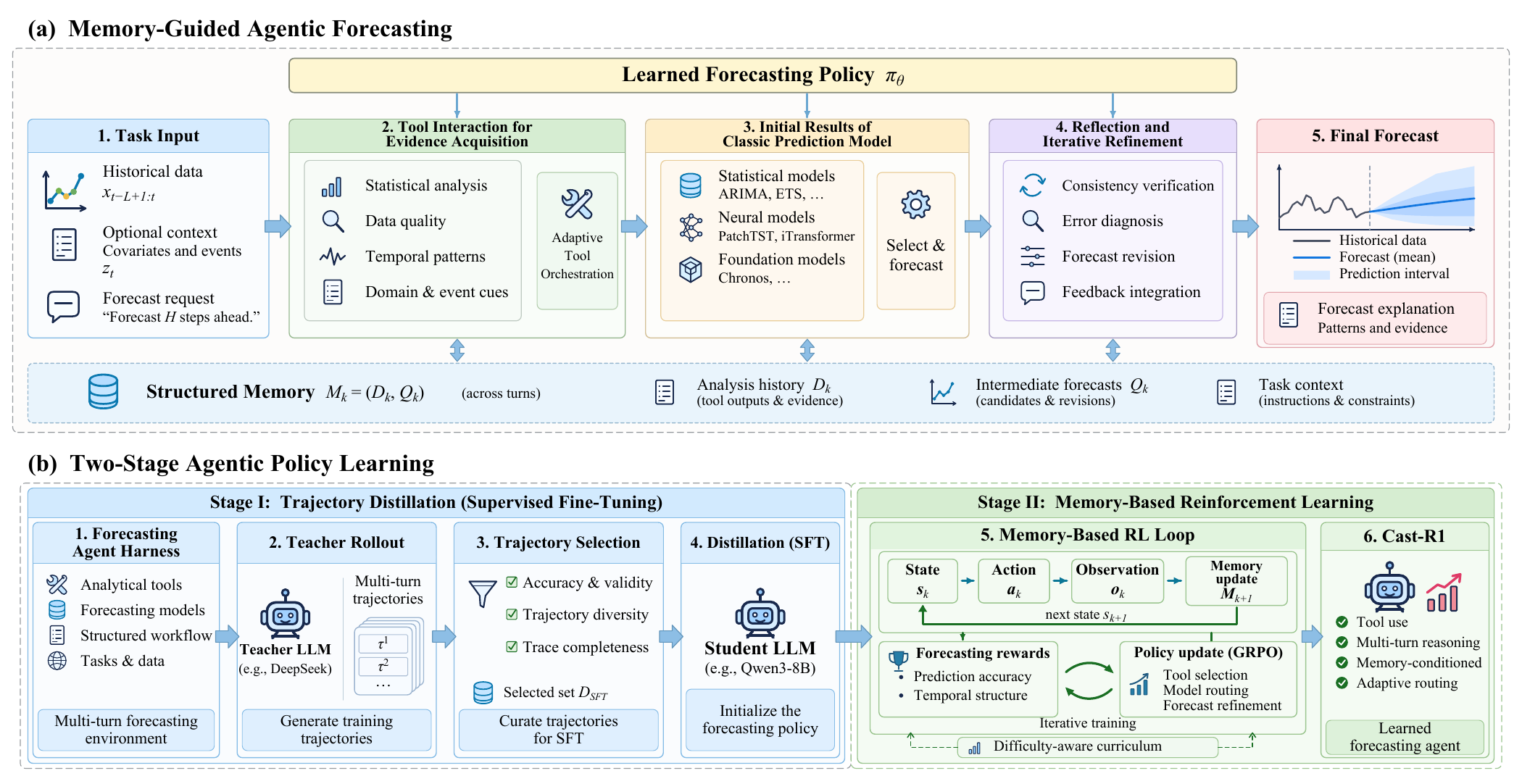}
    \caption{Overview of Cast-R1 as a learned agentic forecasting framework that combines structured memory for state construction with tool-augmented sequential decision-making and policy training.}
    \label{fig:framework}
\end{figure*}

\begingroup
\clubpenalty=10000
\widowpenalty=10000
\displaywidowpenalty=10000
\brokenpenalty=10000
\finalhyphendemerits=100000
\section{The Proposed Cast-R1}
\label{sec:method}
Cast-R1 learns forecasting policies through trajectory distillation and multi-turn GRPO with structured memory. Figure~\ref{fig:framework}(a) presents its memory-guided forecasting workflow; Figure~\ref{fig:framework}(b) summarizes two-stage policy learning.
\subsection{Sequential Forecasting Formulation}
For a look-back length $L$ and forecasting horizon $H$, let $z=(\mathbf{x}_{t-L+1:t},\mathbf{c}_{t-L+1:t},u_t)$ denote the inputs and $y=\mathbf{x}_{t+1:t+H}$ the target. Here, $u_t$ denotes optional task context. For electricity price forecasting, it contains exogenous forecasts or known variables for the prediction window and textual information available at forecast origin $t$; for the long-term tasks, $u_t=\varnothing$. Future target values are excluded from $z$.

We formulate forecasting as a sequence of decision steps $k=1,\ldots,K$, subject to a budget $K\leq K_{\max}$. At step $k$, the agent conditions on the policy-visible state $s_k$ and generates a response $a_k$ with tool calls or the final forecast. Each decision step corresponds to one agent response and may include multiple independent tool calls. Tool calls acquire diagnostic evidence or invoke forecasting models; the final response refines intermediate predictions and outputs the forecast. We denote the trajectory by $\tau=((s_1,a_1),\ldots,(s_K,a_K))$ and its final prediction by $\hat y(\tau)$. An episode ends upon final forecast output or budget exhaustion. Training uses terminal rewards to optimize these decisions, while inference relies on the supplied inputs and \mbox{tool-derived evidence}.

\subsection{Memory-Based State Construction}
\label{subsec:memory_state_management}
Episode-local memory $M_k=(D_k,Q_k)$ retains diagnostic outputs in Analysis History $D_k$ and intermediate forecasts in Prediction Results $Q_k$, starting from $M_1=(\varnothing,\varnothing)$. The prompt $s_k=P(z,M_k)$ combines task inputs and context $z$ with these two memory components.

$P$ selects stage-specific instructions according to the available diagnostics and forecasts. Early prompts include full task inputs; refinement prompts combine recent observations, task context, and memory. Prior agent responses and raw tool messages are not additionally appended as conversation history. For nonterminal decisions:
\begin{equation}
\begin{aligned}
a_k &\sim \pi_\theta(\cdot\mid s_k),\quad o_k=T(s_k,a_k),\\
M_{k+1} &= U(M_k,a_k,o_k),\\
s_{k+1} &= P(z,M_{k+1}),
\end{aligned}
\label{eq:state_transition}
\end{equation}
where $T$ executes tool calls and returns $o_k$. The environment implements $U$ to update memory and $P$ to construct the next prompt from task inputs and retained evidence. Appendix~\ref{app:memory_update} details stage-aware \mbox{prompt construction}.

\begin{figure}[t]
    \centering
    \includegraphics[width=\linewidth]{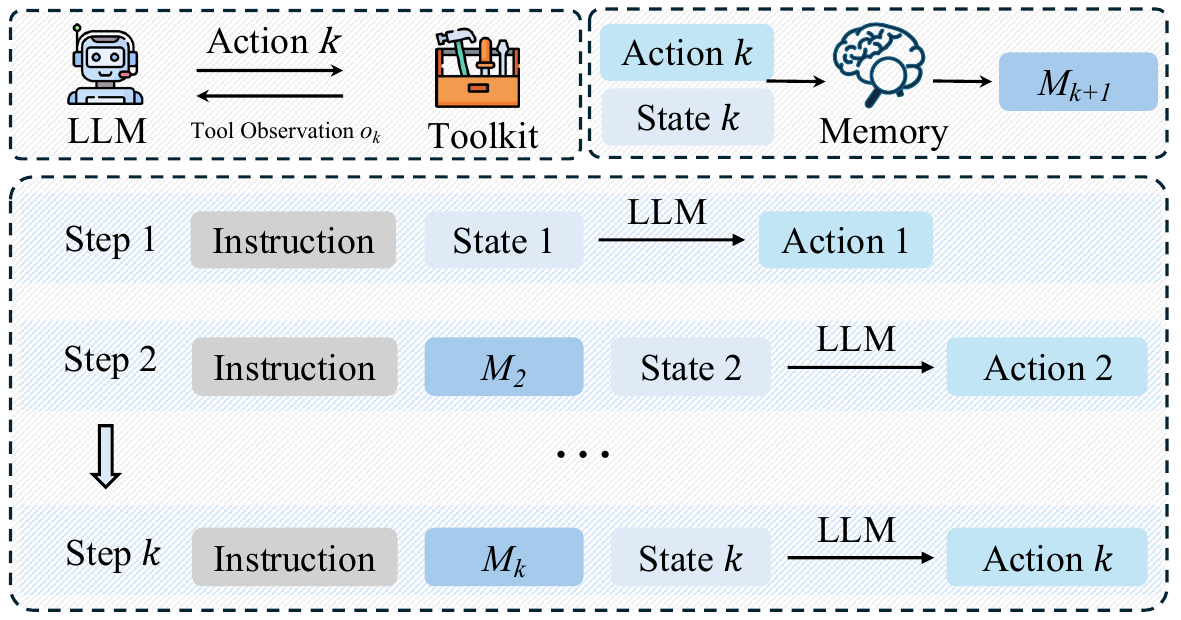}
    \caption{Structured memory for state construction. The environment updates $M_k=(D_k,Q_k)$ with tool observations and constructs the next policy state.}
    \label{fig:memory}
\end{figure}

\subsection{Toolkit Design and Implementation}
\label{subsec:toolkit}
The toolkit supports evidence acquisition and initial forecast generation through two types of \mbox{callable interfaces}.

\paragraph{Diagnostic Tools.}
Five categories cover data quality assessment, global statistical characterization, structural and dynamic analysis, event-level summarization, and model adequacy and residual diagnostics. Their outputs summarize data issues, distributional and frequency-domain properties, regime changes, temporal events, and baseline residuals to support model selection and forecast refinement. Representative statistics and their definitions appear in Appendix~\ref{app:toolkit_statistics}.

\paragraph{Forecasting Model Invocation.}
A unified interface exposes ARIMA, Chronos, PatchTST, and iTransformer as the candidate pool $B$: $\tilde y=f_m(z;H),\quad m\in B$. The policy selects $m$ using accumulated evidence, and the tool returns an $H$-step forecast for subsequent refinement. The pool combines statistical, pretrained, temporal, and \mbox{cross-variable forecasting capabilities}.

\subsection{Memory-Guided Agentic Forecasting}
Figure~\ref{fig:framework}(a) shows three prescribed stages between task input and final output: evidence acquisition (preparation), initial prediction (forecasting), and reflection and refinement. The policy selects tools, routes models, and revises forecasts using $z$ and tool evidence, excluding future targets.

During preparation, diagnostic tools characterize the series and populate Analysis History; independent calls may execute in parallel. During forecasting, the agent selects a model using this evidence and stores its initial forecast in Prediction Results. During refinement, the same policy checks consistency and integrates recent observations, task context, and both memory components to revise the forecast and generate the final output:
\begin{equation}
\hat y=F_\theta(s_{\mathrm{ref}}),
\label{eq:agentic_workflow}
\end{equation}
where $s_{\mathrm{ref}}$ is the refinement state and $F_\theta$ denotes final forecast generation by the same policy~$\pi_\theta$. Stage-aware prompts appear in Appendix~\ref{app:stage_prompt}.

\subsection{Two-Stage Agentic Policy Learning}
Figure~\ref{fig:framework}(b) shows two stages: teacher rollout and trajectory selection supply supervised fine-tuning (SFT); multi-turn GRPO then optimizes the initialized memory-conditioned policy using final-forecast feedback.

\subsubsection{Harness-Guided Trajectory Distillation}
\label{subsec:trajectory_distillation}
DeepSeek-V3.2~\cite{deepseekai2025v32} generates multi-turn demonstrations of diagnostic tool use, model selection, and forecast refinement in the forecasting harness, using only training-split windows from the ten datasets. Retained trajectories train the Qwen3-8B student through state--response supervision conditioned on accumulated tool feedback.

\paragraph{Trajectory Evaluation and Selection.}
We apply execution-validity filtering followed by forecasting-quality selection. Incomplete trajectories and those with failed tools, malformed outputs, or non-finite predictions are discarded. Valid trajectories are ranked by ascending final-forecast MSE within each dataset using training targets. We retain the 120 lowest-error trajectories per dataset, yielding 1,200 complete demonstrations in $D_{\mathrm{SFT}}$ with balanced dataset representation. Appendix~\ref{app:sft_protocol} details the completeness checks, allocation for dataset diversity, and SFT settings.
Trajectory distillation minimizes the following SFT objective over demonstration trajectories $\tau^\star\sim D_{\mathrm{SFT}}$:
\begin{equation}
L_{\mathrm{SFT}}(\theta)
=-\mathbb{E}_{\tau^\star}
\left[\sum_k\log\pi_\theta(a_k^\star\mid s_k^\star)\right].
\label{eq:sft_objective}
\end{equation}
Here, $s_k^\star$ is the demonstrated state, including tool feedback, and $a_k^\star$ is the teacher response containing tool calls or the final forecast. Only teacher responses are supervised.

\subsubsection{Multi-turn RL with Structured Memory}
\label{subsec:reinforcement_learning}
We apply group relative policy optimization (GRPO)~\cite{shao2024deepseekmath} to the distilled policy in a multi-turn forecasting environment~\cite{wang2025ragen}. The environment updates memory and constructs states as in Equation~\ref{eq:state_transition}, while GRPO updates the shared policy $\pi_\theta$. Earlier tool choices therefore shape the evidence available to subsequent decisions. Relative advantages are estimated from terminal rewards of trajectories sampled for the same input, without a separate critic.
The  objective is to maximize expected terminal reward:
\begin{equation}
J(\theta)=\mathbb{E}\bigl[R(\tau,y)\bigr],
\label{eq:rl_objective}
\end{equation}
where the expectation is over training input--target pairs and policy rollouts, and $R(\tau,y)$ evaluates the final forecast.
For each input, GRPO samples $G$ complete trajectories $\{\tau_i\}_{i=1}^{G}$ using the rollout policy $\pi_{\mathrm{old}}$. With $R_i=R(\tau_i,y)$, the group-relative advantage is:
\begin{equation}
\hat A_i=\frac{R_i-\bar R}{\sigma_R+\delta},
\label{eq:group_advantage}
\end{equation}
where $\bar R$ and $\sigma_R$ are the group mean and standard deviation, and $\delta>0$ is a numerical stabilizer. The same trajectory-level advantage supervises agent-generated tokens across its turns, while tool outputs serve as environment observations. GRPO uses these advantages in clipped policy updates with KL regularization against a reference policy. Appendix~\ref{app:rl_protocol} gives the token-level formulation; Table~\ref{tab:hyperparameters} lists the training settings.

\paragraph{Multi-view Reward Design.}
The terminal reward combines numerical accuracy, temporal structure, and output completeness to guide forecasting decisions. For outputs with valid \texttt{<answer>} and \texttt{</answer>} tags, the reward is:
\begin{equation}
R_{\mathrm{valid}}=p\left(w_{\mathrm{len}}+\sum_j w_j u_j\right),
\label{eq:forecast_reward}
\end{equation}
where $p=\min(1,n_{\mathrm{pred}}/n_{\mathrm{gt}})$ uses the prediction and target lengths $n_{\mathrm{pred}}$ and $n_{\mathrm{gt}}$. The sum ranges over $j\in\{\mathrm{pred},\mathrm{seas},\mathrm{trend},\max,\min\}$, covering prediction accuracy, seasonality, trend, and maximum/minimum alignment. Prediction, seasonal, and trend MSEs are mapped to scores in $[0,1]$ by $q(e)=[1+\log(1+e)]^{-1}$ for $e\geq0$; alignment terms use extremum hit rates. Missing answer tags yield $R=-1$. Weights and scoring details appear in Appendix~\ref{app:reward_details}.
Rewards are computed exclusively on training-split samples using their targets; validation and test targets are excluded.

\paragraph{Curriculum Learning.}
The difficulty-aware curriculum in Figure~\ref{fig:framework}(b) organizes training samples along two dimensions: prediction MSE of a reference forecaster and data-intrinsic complexity measured by normalized permutation entropy~\cite{bandt2002permutation}. The curriculum filters noise-dominated samples rather than selecting examples solely by high prediction error, and progressively introduces retained samples from simple trend and seasonal patterns to more complex temporal structures. Appendix~\ref{app:curriculum} defines the entropy statistic and distinguishes the reference forecaster from the trajectory teacher.
\par
\endgroup

\section{Experiments}

\begin{table*}[t]
	\centering
	\footnotesize
	\setlength{\tabcolsep}{1pt}
	\newcommand{\mseresultwidth}{\dimexpr(\linewidth-82pt-22\tabcolsep-2\arrayrulewidth)/10\relax}
	\begin{tabular}{>{\centering\arraybackslash}m{82pt}|*{5}{>{\centering\arraybackslash}m{\mseresultwidth}}|*{5}{>{\centering\arraybackslash}m{\mseresultwidth}}}
		\toprule
		& \multicolumn{5}{c|}{\shortstack{\textbf{LTSF (Historical Inputs)}\\($L=96$, $H=96$)}} & \multicolumn{5}{c}{\shortstack{\textbf{EPF (Forecast-Origin Context)}\\($L=168$, $H=24$)}} \\
		\cmidrule(lr){2-6}\cmidrule(lr){7-11}
		Model & \multicolumn{1}{c}{ETTh1} & \multicolumn{1}{c}{ETTh2} & \multicolumn{1}{c}{ETTm1} & \multicolumn{1}{c}{ETTm2} & \multicolumn{1}{c|}{Wind} & \multicolumn{1}{c}{BE} & \multicolumn{1}{c}{DE} & \multicolumn{1}{c}{FR} & \multicolumn{1}{c}{NP} & \multicolumn{1}{c}{PJM} \\
		\midrule
		ARIMA & 10.879 & 12.567 & 9.898 & 22.527 & 2166.64 & 869.444 & 362.487 & 963.997 & 55.507 & 149.281 \\
		Prophet & 46.697 & 23.525 & 12.225 & 31.525 & 8071.24 & 749.864 & 340.371 & 1024.401 & 40.754 & 53.777 \\
		PatchTST & 9.300 & 11.358 & 5.838 & 16.386 & 2024.83 & 512.834 & 201.705 & 807.512 & 29.428 & 27.164 \\
		iTransformer & \underline{7.505} & 10.161 & 5.771 & 12.751 & 1591.64 & 645.760 & 265.310 & 824.473 & 34.591 & 53.751 \\
		ConvTimeNet & 8.032 & 18.072 & 6.358 & \underline{12.253} & 1660.13 & 668.275 & \underline{198.431} & 789.035 & 29.903 & 34.638 \\
		TimeXer & 8.765 & 13.798 & 4.387 & 14.136 & 2063.99 & 772.704 & 222.798 & 803.202 & 32.052 & 34.283 \\
		DLinear & 8.506 & 12.347 & 5.525 & 15.272 & 2169.82 & 739.149 & 239.753 & \underline{787.498} & 37.287 & 40.562 \\
			Chronos & 9.397 & 16.729 & 4.414 & 24.440 & 1764.15 & \underline{497.075} & 211.205 & 812.732 & 30.804 & 29.697 \\
		TimesFM & 8.096 & \underline{9.932} & 5.996 & 15.232 & 1445.52 & 502.345 & 212.005 & 852.003 & 30.557 & 30.978 \\
		OFA & 14.528 & 26.227 & 10.024 & 21.551 & 2044.332 & 621.548 & 275.331 & 1004.898 & 43.227 & 44.669 \\
		Time-LLM & 10.523 & 15.527 & 5.137 & 16.001 & 1524.75 & 523.552 & 232.875 & 825.667 & 35.486 & 31.998 \\
		TimeReasoner & 7.965 & 11.212 & 6.864 & 17.003 & 1662.55 & 549.797 & 286.667 & 903.977 & 46.998 & 35.778 \\
		Time-R1 & 8.989 & 13.548 & \underline{4.238} & 15.995 & \underline{1353.938} & 535.372 & 279.146 & 880.259 & 45.765 & 34.839 \\
		\mbox{TimeSeriesScientist} & 23.207 & 30.909 & 8.278 & 17.445 & 3400.916 & 906.681 & 245.400 & 904.655 & 46.313 & 45.400  \\
		AlphaCast & 13.395 & 17.350 & 6.742 & 15.185 & 1367.628 & 653.983 & 248.981 & 804.892 & \underline{27.113} & \textbf{26.535}  \\
		\midrule
		\rowcolor[gray]{0.95} \textbf{Cast-R1} & \textbf{6.062} & \textbf{8.405} & \textbf{3.465} & \textbf{11.181} & \textbf{1331.513} & \textbf{473.427} & \textbf{183.781} & \textbf{776.898} & \textbf{24.750} & \underline{26.905} \\
		\bottomrule
	\end{tabular}
			\caption{MSE results for two benchmark families, not a controlled context ablation. EPF provides forecast-origin context; input access follows Appendix~\ref{app:input_access}. Lower values are better. Best results are \textbf{bold}; second-best results are \underline{underlined}. MAE results appear in Appendix Table~\ref{tab:main_results_mae}.}
	\label{tab:main_results}
\end{table*}

\subsection{Experimental Setup}
\paragraph{Datasets.}
We evaluate four ETT datasets~\cite{zhou2021informer} and Wind~\cite{qiu2024tfb} for LTSF, and five regional EPF datasets (NP, PJM, BE, FR, and DE)~\cite{lago2021forecasting}. Table~\ref{tab:dataset} reports their sizes, variables, and temporal resolutions. LTSF uses historical targets and covariates; EPF additionally provides forecast-horizon exogenous variables and dataset-provided text available at the forecast origin. These benchmark families do not constitute a controlled context ablation. Dataset details appear in Appendix~\ref{app:datasets}.
\begin{table}[t]
\centering
\footnotesize
\setlength{\tabcolsep}{4pt}
\begin{tabular*}{\linewidth}{@{\extracolsep{\fill}}lrrl}
\toprule
Dataset & Length & Variables & Frequency \\
\midrule
ETTh1 \& ETTh2 & 17,420 & 7 & 1 hour \\
ETTm1 \& ETTm2 & 69,680 & 7 & 15 mins \\
Wind & 48,673 & 7 & 15 mins \\
BE & 14,496 & 3 & 1 hour \\
DE & 14,496 & 4 & 1 hour \\
FR & 14,496 & 3 & 1 hour \\
NP & 14,496 & 3 & 1 hour \\
PJM & 14,496 & 3 & 1 hour \\
\bottomrule
\end{tabular*}
\caption{Statistics of the benchmark datasets, including length, variables, and sampling frequency. }
\label{tab:dataset}
\end{table}

\begin{table*}[t]
\centering
\footnotesize
\setlength{\tabcolsep}{2pt} 
\renewcommand{\arraystretch}{1.15}

\resizebox{\linewidth}{!}{%
\begin{tabular}{c|cc|cc|cc|cc|cc|cc}
\toprule
\multirow{2}{*}{\textbf{Variant}} & \multicolumn{2}{c|}{ETTh1} & \multicolumn{2}{c|}{ETTm1} & \multicolumn{2}{c|}{Wind} & \multicolumn{2}{c|}{DE} & \multicolumn{2}{c|}{NP} & \multicolumn{2}{c}{PJM} \\
 & MSE & MAE & MSE & MAE & MSE & MAE & MSE & MAE & MSE & MAE & MSE & MAE \\ \midrule

w/o Features & 7.145 & 1.988 & 4.979 & 1.548 & 1743.431 & 23.122 & 194.672 & 9.864 & 28.714 & 4.260 & 30.641 & 4.251 \\

w/o Prediction & 15.993 & 2.144 & 13.134 & 2.258 & 4047.163 & 28.113 & 564.157 & 12.965 & 64.813 & 5.267 & 91.122 & 7.135 \\  \midrule
\rowcolor[gray]{0.95} \textbf{Cast-R1 } & \textbf{6.062} & \textbf{1.320} & \textbf{3.465} & \textbf{1.160} & \textbf{1331.513} & \textbf{16.125} & \textbf{183.781} & \textbf{7.400} & \textbf{24.750} & \textbf{3.255} & \textbf{26.905} & \textbf{3.877} \\
\bottomrule
\end{tabular}%
}
\caption{Toolkit ablations on representative datasets. Removing feature extraction or model prediction increases error. Best results are \textbf{bold}.}
\label{tab:ablation_components}
\end{table*}

\paragraph{Baselines.}
We compare \model{} with fifteen baselines across five categories.
Statistical methods include ARIMA~\cite{hyndman2008automatic} and Prophet~\cite{taylor2018forecasting}.
Deep learning methods include PatchTST~\cite{Yuqietal-2023-PatchTST}, iTransformer~\cite{liuitransformer}, TimeXer~\cite{wang2024timexer}, ConvTimeNet~\cite{cheng2025convtimenet}, and DLinear~\cite{zeng2023transformers}.
Foundation models include Chronos~\cite{ansari2025chronos} and TimesFM~\cite{das2024decoder}.
LLM-based forecasters include OFA~\cite{zhou2023one}, Time-LLM~\cite{jintime}, TimeReasoner~\cite{cheng2025can}, and Time-R1~\cite{luo2025time}. Time-R1 trains slow-thinking temporal reasoning via supervised warm-up and reinforcement learning.
Agentic TSF frameworks include AlphaCast~\cite{zhang2025alphacast} and TimeSeriesScientist~\cite{zhao2025timeseriesscientist}, which coordinate forecasting through LLM-driven analysis and operations.
Baseline details are provided in Appendix~\ref{app:baselines}; EPF input access follows Appendix~\ref{app:input_access}.

\paragraph{Implementation Details.}
We use Qwen3-8B~\cite{yang2025qwen3} as the backbone and train a single agent on all ten datasets with four NVIDIA A800 GPUs. Trajectory distillation uses SFT on 1,200 selected teacher trajectories (120 per dataset) for 1 epoch with a learning rate of $1.0\times10^{-5}$, followed by GRPO-based RL with learning rate $1.0\times10^{-6}$, batch size 64, group size $G=8$, temperature 1.0, and maximum prompt/response lengths of 8192/4096 tokens. Baselines follow official settings. All methods use matched input windows and horizons. We evaluate MSE and MAE with chronological splits of 6:2:2 (ETT) and 7:1:2 (others). We exclude future target values from the inputs, and RL rewards use only training samples.
Full settings appear in Appendix~\ref{app:implementation}.

\subsection{Main Results}
Table~\ref{tab:main_results} reports MSE results for the ten datasets. \model{} achieves the lowest MSE on nine datasets and the second-lowest on PJM, where AlphaCast performs better. Averaging the per-dataset relative MSE reductions against the lowest baseline MSE, including the negative reduction on PJM, gives 8.41\% overall, 12.65\% for the five long-term datasets, and 4.16\% for the five electricity price datasets. Compared with the agentic TSF baselines, \model{} obtains lower MSE and MAE than TimeSeriesScientist on all ten datasets, and lower MSE than AlphaCast on nine and lower MAE on seven. Against the trained reasoning model Time-R1, \model{} obtains lower MSE on all ten datasets and lower MAE on eight. These comparisons assess the complete framework against both agentic forecasting and forecasting-specific LLM training. Appendix Table~\ref{tab:main_results_mae} reports MAE results. \model{} achieves the lowest MAE on seven datasets; AlphaCast leads on Wind and PJM, while Chronos leads on NP. The difference on Wind, where \model{} has the lowest MSE but the third-lowest MAE, illustrates metric-dependent rankings. The following ablations examine how removing individual components affects the complete framework.
\begin{figure*}[t]
    \centering
    \includegraphics[width=\linewidth]{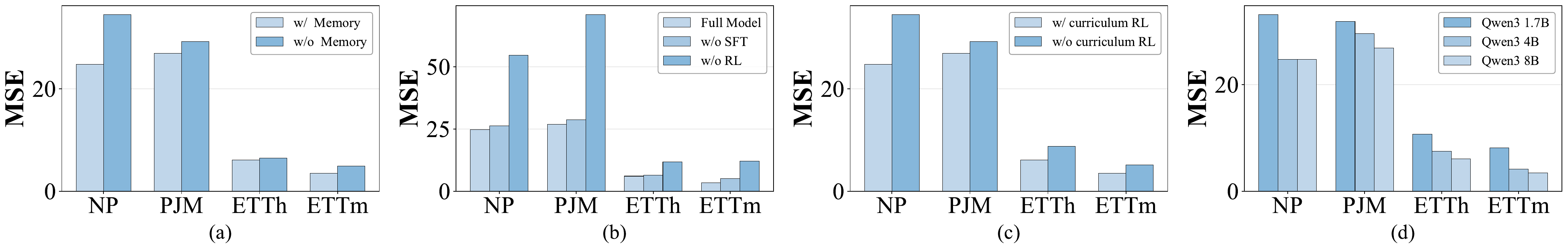}
    \caption{MSE comparisons for (a) memory-based state construction, (b) trajectory distillation (SFT) and RL, (c) curriculum learning, and (d) backbone size. ETTh/ETTm denote ETTh1/ETTm1.}
    \label{fig:all_mse}
\end{figure*}
\begin{figure*}[t]
    \centering
    \includegraphics[width=\linewidth]{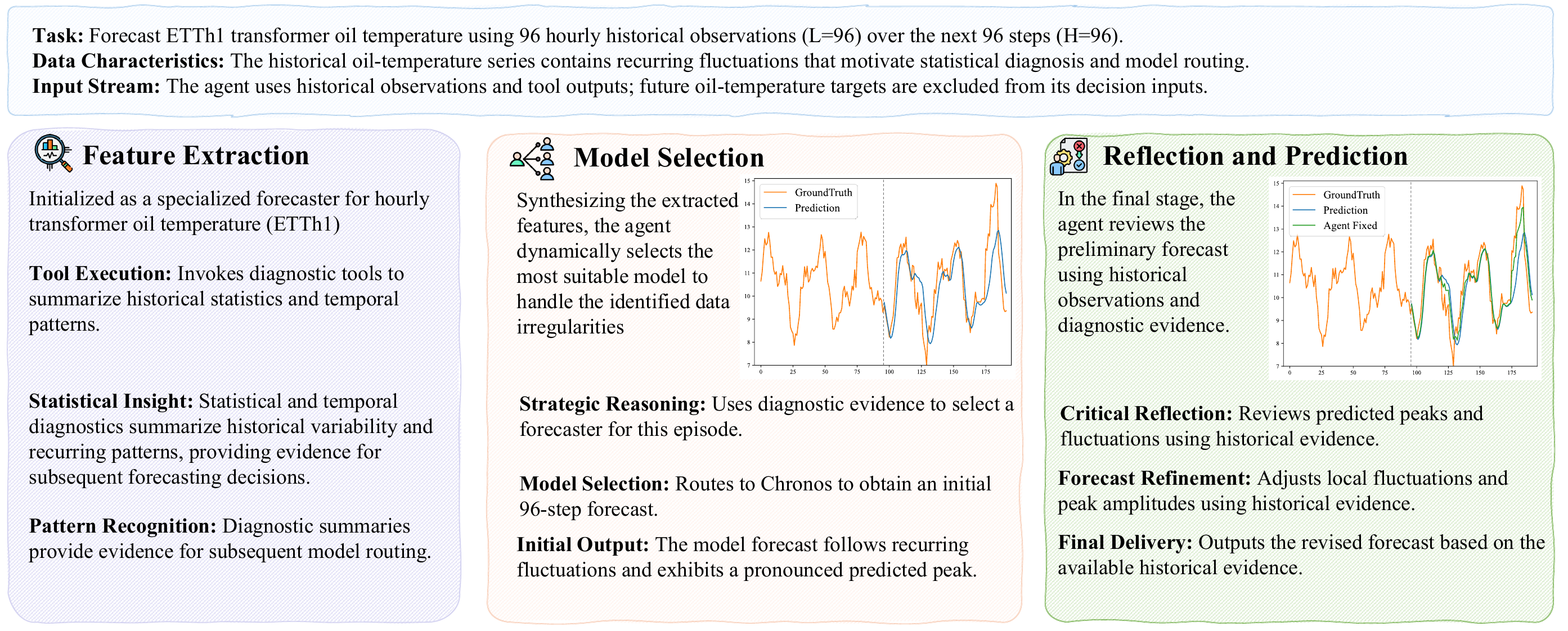}
    \caption{Qualitative ETTh1 oil-temperature forecasting case with $L=96$ and $H=96$, illustrating statistical diagnosis, routing to Chronos, and forecast refinement. Dashed lines mark the history--forecast boundary.}
    \label{fig:case_study}
\end{figure*}

\subsection{Training Ablations}
\subsubsection{Trajectory Distillation and RL Optimization}
Figure~\ref{fig:all_mse}(b) compares forecasting errors, while Figure~\ref{fig:training_dynamics}(a) reports training rewards. Trajectory distillation refers to the SFT stage in Section~\ref{subsec:trajectory_distillation}. The complete pipeline has the lowest MSE on the four displayed benchmarks; removing RL yields the largest error increase, and omitting distillation (w/o SFT) also increases error. The run initialized by distillation starts with a higher reward than pure RL and generally has higher rewards during training. A possible explanation is that the teacher trajectories provide a starting policy for subsequent optimization. These observations favor combining trajectory distillation and multi-turn RL.

\begin{table}[t]
\centering

\footnotesize
\renewcommand{\arraystretch}{1.2}
\setlength{\tabcolsep}{0.8pt}
\begin{tabular*}{\linewidth}{@{\extracolsep{\fill}}l|rr|rr|rr}
\toprule
\multirow{2}{*}{\textbf{Variant}} & \multicolumn{2}{c|}{ETTh1} & \multicolumn{2}{c|}{ETTm1} & \multicolumn{2}{c}{Wind} \\
 & MSE & MAE & MSE & MAE & MSE & MAE \\ \midrule

w/o Length & 6.102 & 1.928 & 3.982 & 1.378 & 1575.997 & 22.445 \\
w/o Error & 13.444 & 2.832 & 12.731 & 2.681 & 3146.645 & 32.329 \\
w/o Trend & 8.452 & 2.347 & 6.684 & 1.614 & 1830.842 & 25.219 \\
w/o Align. & 7.847 & 2.031 & 5.984 & 1.483 & 1744.896 & 24.786 \\ \midrule
\rowcolor[gray]{0.95} \textbf{Cast-R1} & \textbf{6.062} & \textbf{1.320} & \textbf{3.465} & \textbf{1.160} & \textbf{1331.513} & \textbf{16.125} \\
\bottomrule
\end{tabular*}
\caption{Reward ablations, where Length denotes length weighting, Error measures prediction accuracy, Trend captures both trend and seasonality effects, and Align represents alignment. }
\label{tab:reward_ablation}
\end{table}

\subsubsection{Reward Component Contributions}
Table~\ref{tab:reward_ablation} ablates the reward groups in Equation~\ref{eq:forecast_reward}: length weighting, prediction accuracy, trend and seasonality, and maximum/minimum alignment. The complete reward consistently achieves the lowest reported MSE and MAE, and removing the prediction error term yields the largest MSE increase on all three datasets. Removing the other groups also increases the reported errors, although the observed ETTh1 MSE increase without length weighting is only 0.66\%. Structural terms may supply complementary optimization signals; these error metrics do not directly measure preservation of temporal patterns or output completeness.

\subsubsection{Curriculum-Guided Training}
Figure~\ref{fig:all_mse}(c) shows lower reported MSE with the curriculum on the four displayed benchmarks. These observed improvements favor the combination of sample filtering and progressive ordering. Exposure to increasingly difficult samples is a possible contributor, alongside changes in the retained training data. Separating filtering from ordering requires comparisons with identical retained samples and matched update budgets; the current results assess their combined effect.
\subsection{Agent Ablations}
\subsubsection{Memory-Based State Construction}
Figure~\ref{fig:all_mse}(a) compares the complete framework with its without-memory variant. The complete framework has lower reported MSE on the four displayed benchmarks, favoring its memory-based state configuration within the evaluated workflow. The comparison with memory removed does not isolate the effect of structured state organization from retaining full interaction history. These results assess forecasting performance; they do not directly measure tool-selection correctness or reasoning reliability.
The stage-aware prompt construction and memory update strategy are further described in the Appendix~\ref{app:memory_update}.

\begin{figure}[!t]
    \centering
    \includegraphics[width=1\linewidth]{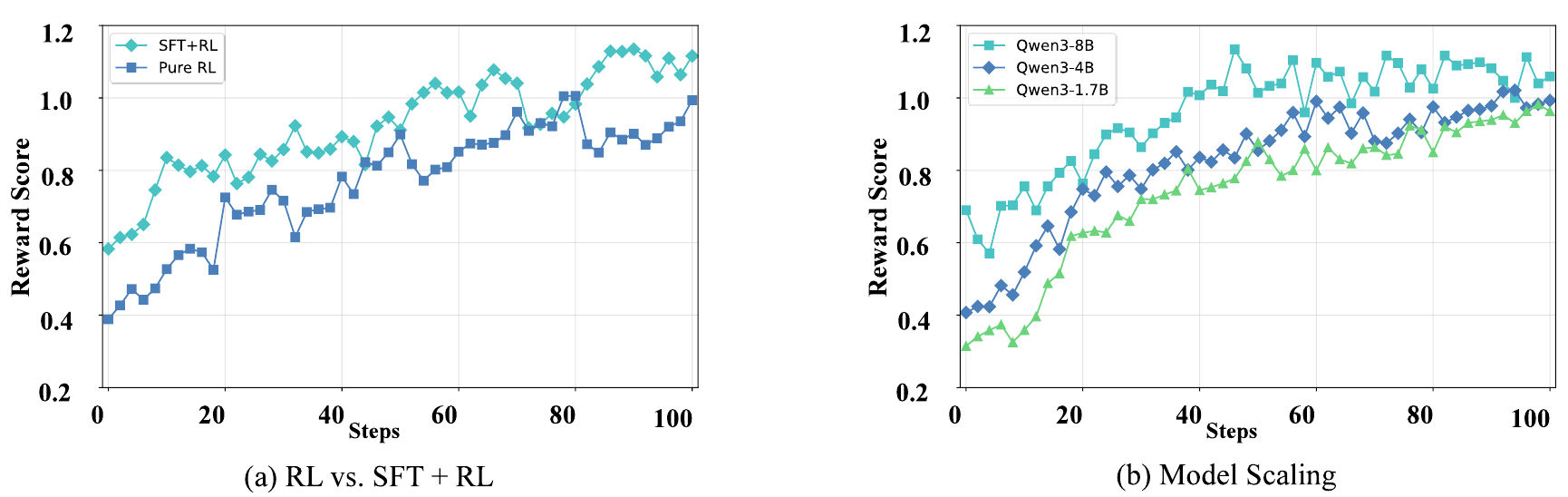}
    \caption{Training reward curves comparing (a) initialization by trajectory distillation (SFT) with pure RL and (b) different backbones.}
    \label{fig:training_dynamics}
\end{figure}

\subsubsection{Diagnostic and Prediction Tools}
Table~\ref{tab:ablation_components} reports ablations of diagnostic feature extraction (w/o Features) and forecasting model invocation (w/o Prediction).
Removing either component increases MSE and MAE across all six datasets compared with the full framework, with larger performance degradation generally observed when prediction tools are removed.
Appendix Tables~\ref{tab:feature_tools} and~\ref{tab:model_tools} further examine individual tool contributions.
While most tool removals degrade forecasting performance, some yield improvements on specific datasets or metrics.
These results suggest that component-level contributions are relatively consistent, whereas the effects of individual tools vary across datasets and evaluation metrics.

\subsection{Backbone Scaling Analysis}
Figure~\ref{fig:all_mse}(d) compares forecasting performance across Qwen3 backbone sizes, while Figure~\ref{fig:training_dynamics}(b) presents their training rewards. Larger backbones achieve lower reported MSE on several benchmarks, although improvements are not uniform across all settings. The largest evaluated backbone attains the lowest or tied-lowest final MSE and obtains higher training rewards than the smallest variant. Together, these results suggest that increased model capacity can benefit both optimization and forecasting performance within the evaluated Qwen3 family, without implying consistent gains beyond the tested configurations. Moreover, the differences across benchmarks suggest that scaling effects depend on dataset characteristics.

\subsection{Case Study}
Figure~\ref{fig:case_study} shows an ETTh1 oil-temperature forecasting episode with 96 hourly observations and a 96-step horizon. The agent uses statistical and temporal diagnostics to select Chronos based on evidence. The initial forecast captures recurring oscillations but shows local timing and amplitude errors. During reflection, the agent compares these deviations with historical evidence and adjusts selected fluctuations and peaks. This illustrates interactions among diagnostics, predictions, and reflection. Decisions rely solely on supplied inputs and tool outputs, without future oil-temperature targets. As a qualitative example, it emphasizes the decision trace rather than consistent improvement. 

\section{Conclusion}
We present Cast-R1, an 8B-parameter agentic time series forecasting model trained through trajectory distillation and multi-turn GRPO with structured memory.
Teacher trajectories initialize forecasting-oriented tool use, while GRPO optimizes tool selection, model routing, and forecast refinement using accuracy and temporal-structure rewards.
Episode-local memory supplies diagnostic evidence and intermediate forecasts to subsequent decisions.
Cast-R1 achieved the lowest MSE on most of the ten benchmark datasets. Ablation studies further indicated that multi-stage training and structured memory contributed to its forecasting performance.

\section*{Limitations}
A single agent is trained on all ten datasets using Qwen3 backbones. The evaluation therefore addresses performance within these benchmark domains; transfer to unseen datasets, other domains, and other language-model families remains to be established. The harness prescribes the stage order and candidate forecasting tools, and the results do not establish open-ended workflow planning or robustness to changes in the tool pool. Memory retains evidence within each episode; cross-task experience retrieval is outside the evaluated setting.

MSE and MAE assess numerical forecasts but do not directly evaluate the faithfulness of the accompanying natural-language reasoning. The reported comparisons provide point estimates, and small differences require uncertainty estimates across independent runs. We also do not systematically measure the inference cost and latency of the multi-step agent workflow relative to single-model forecasting baselines.

\bibliography{kdd2026}

\clearpage
\appendix
\section{Datasets and Evaluation Protocol}
\label{app:datasets}
We evaluate ten datasets: five short-term electricity price forecasting (EPF) benchmarks and five long-term time series forecasting (LTSF) benchmarks. Each task predicts a single target series from historical observations and available covariates. Table~\ref{tab:datasets_detail} lists the targets and original covariate fields; the following subsections describe the datasets, task configuration, input access, and evaluation metrics.

\begin{table*}[!t]
    \centering
    \caption{Forecast targets and original covariate fields. EPF additionally uses forecast-horizon covariates and textual context available at the forecast origin. LTSF uses covariates only within the historical window. Textual context is additional to the listed fields.}
    \label{tab:datasets_detail}
    \footnotesize
    \renewcommand{\arraystretch}{1.15}
    \setlength{\tabcolsep}{4pt}
    \begin{tabular}{@{}p{0.18\textwidth}p{0.14\textwidth}p{0.08\textwidth}p{0.06\textwidth}p{0.46\textwidth}@{}}
    \toprule
    \textbf{Dataset} & \textbf{Target} & \textbf{Freq.} & \textbf{Task} & \textbf{Covariate fields} \\
    \midrule
    NP & Electricity price & 1 h & EPF & Grid load and wind generation forecasts. \\
    PJM & Electricity price & 1 h & EPF & System-wide and zonal load forecasts. \\
    BE & Electricity price & 1 h & EPF & National load and French generation forecasts. \\
    FR & Electricity price & 1 h & EPF & Grid load and domestic generation forecasts. \\
    DE & Electricity price & 1 h & EPF & Zonal load, wind, and solar generation forecasts. \\
    \midrule
    ETTh1/ETTh2 & Oil temperature & 1 h & LTSF & Six power load variables accompanying oil temperature. \\
    ETTm1/ETTm2 & Oil temperature & 15 min & LTSF & The same six power load variables as the hourly ETT variants. \\
    Wind & Wind power & 15 min & LTSF & Weather forecasts and measured wind speed. \\
    \bottomrule
    \end{tabular}
\end{table*}

\subsection{Electricity Price Forecasting}
The EPF benchmarks~\cite{lago2021forecasting} contain hourly day-ahead electricity prices from five markets: Nordic Nord Pool (NP), the Commonwealth Edison (ComEd) zone of PJM, Belgium (BE), France (FR), and Germany (DE). Their load and generation forecast fields are listed in Table~\ref{tab:datasets_detail}; the German load forecast covers the Amprion area. EPF additionally supplies forecast-horizon covariates and dataset-provided text available at the forecast origin, following the input-access protocol in Appendix~\ref{app:input_access}.

\subsection{Long-term Forecasting}
\paragraph{ETT.} The Electricity Transformer Temperature benchmark~\cite{zhou2021informer} records oil temperature (OT) and six load variables (HUFL, HULL, MUFL, MULL, LUFL, and LULL) over two years. OT is the target forecasting variable in this study. ETTh1 and ETTh2 use hourly sampling; ETTm1 and ETTm2 use 15-minute sampling.

\paragraph{Wind.} The Wind dataset from TFB~\cite{qiu2024tfb} contains 48,673 observations at 15-minute intervals and seven variables. Wind power is the target; the accompanying fields are forecasts of wind speed, wind direction, temperature, pressure, and humidity, plus measured wind speed. These fields are used only within the historical window.

\subsection{Forecasting Task Configuration}
\label{app:task_configuration}
The chronological train/validation/test split is 6:2:2 for ETT and 7:1:2 for the other datasets. RL rewards use training-split targets only. Table~\ref{tab:task_protocol} gives the matched input windows and horizons. A horizon of 96 steps corresponds to 96 hours for ETTh and 24 hours for the 15-minute ETTm and Wind series.

\begin{table}[!htbp]
    \centering
    \caption{Forecasting task configuration. $L$ and $H$ count time steps; duration gives the physical length of the forecasting horizon. Split denotes chronological train/validation/test proportions.}
    \label{tab:task_protocol}
    \footnotesize
    \setlength{\tabcolsep}{3pt}
    \begin{tabular*}{\linewidth}{@{\extracolsep{\fill}}lrrll}
    \toprule
    \textbf{Family} & $L$ & $H$ & \textbf{Duration} & \textbf{Split} \\
    \midrule
    EPF & 168 & 24 & 24 hours & 7:1:2 \\
    ETTh1/ETTh2 & 96 & 96 & 96 hours & 6:2:2 \\
    ETTm1/ETTm2 & 96 & 96 & 24 hours & 6:2:2 \\
    Wind & 96 & 96 & 24 hours & 7:1:2 \\
    \bottomrule
    \end{tabular*}
\end{table}
\subsection{Input Access}
\label{app:input_access}
All methods receive historical target observations. Covariate-capable methods receive the same dataset-provided variables within the historical window. LTSF uses these historical inputs. For EPF, \model{} additionally uses forecast-horizon exogenous covariates and dataset-provided textual context available at the forecast origin. Baselines use these additional inputs when supported by their input interfaces; they do not consume unsupported modalities. Future target values are excluded from the inputs in both benchmark families. Tables~\ref{tab:main_results} and~\ref{tab:main_results_mae} therefore compare complete forecasting systems under their supported input interfaces. Matched windows and horizons do not imply identical input modalities, and these comparisons do not isolate the effect of additional context.

\subsection{Evaluation Metrics}
\label{app:evaluation_metrics}
We report mean squared error (MSE) and mean absolute error (MAE) on the raw data scale:
\begin{equation}
\begin{aligned}
\mathrm{MSE} &= \frac{1}{H}\sum_{i=1}^{H}(y_i-\hat y_i)^2,\\
\mathrm{MAE} &= \frac{1}{H}\sum_{i=1}^{H}|y_i-\hat y_i|.
\end{aligned}
\end{equation}
Here, $y_i$ and $\hat y_i$ are the target and prediction at horizon step $i$; these equations define the errors for one forecasting episode. We evaluate forecasting accuracy over the complete prediction horizon, capturing both large deviations through MSE and overall prediction discrepancies through MAE. Lower values indicate better forecasting performance under both metrics.

\paragraph{Relative MSE Reduction.}
The improvement reported in the main text averages relative reductions within datasets. Let $b_d$ be the lowest baseline MSE and $c_d$ the Cast-R1 MSE on dataset $d$. For a dataset group $\mathcal{D}$, the reported percentage is:
\begin{equation}
\Delta_{\mathcal{D}}=\frac{100}{|\mathcal{D}|}\sum_{d\in\mathcal{D}}\frac{b_d-c_d}{b_d}.
\end{equation}
Each dataset contributes equally, and negative reductions are retained. This gives 8.41\% across all ten datasets, 12.65\% for LTSF, and 4.16\% for EPF. The calculation uses relative errors because raw-scale MSEs have different units across datasets. Accordingly, the reported gain should be interpreted as the average percentage improvement across datasets, rather than the reduction obtained by pooling squared errors over all available individual forecasting test observations.

\FloatBarrier
\section{Baselines}
\label{app:baselines}
We compare \model{} with fifteen baselines in five categories. Shared evaluation settings appear in Appendix~\ref{app:task_configuration}, and implementation settings are summarized below in Appendix~\ref{app:baseline_implementation}.

\subsection{Baseline Methods}
\paragraph{Statistical Methods.}
\begin{itemize}[leftmargin=*, noitemsep, topsep=2pt]
    \item \textbf{ARIMA}~\cite{hyndman2008automatic}: autoregression, differencing, and moving averages for statistical extrapolation.
    \item \textbf{Prophet}~\cite{taylor2018forecasting}: an additive model of trend, seasonality, and holiday effects.
\end{itemize}

\paragraph{Deep Learning Methods.}
\begin{itemize}[leftmargin=*, noitemsep, topsep=2pt]
    \item \textbf{DLinear}~\cite{zeng2023transformers}: linear forecasting with trend and seasonal decomposition.
    \item \textbf{PatchTST}~\cite{Yuqietal-2023-PatchTST}: channel-independent temporal modeling with patch-based Transformer inputs.
    \item \textbf{iTransformer}~\cite{liuitransformer}: attention across variate tokens to model cross-variable dependencies.
    \item \textbf{TimeXer}~\cite{wang2024timexer}: forecasting that incorporates and aligns exogenous variables.
    \item \textbf{ConvTimeNet}~\cite{cheng2025convtimenet}: hierarchical convolutional forecasting with adaptive segmentation and deformable patching.
\end{itemize}

\paragraph{Foundation Models.}
\begin{itemize}[leftmargin=*, noitemsep, topsep=2pt]
    \item \textbf{Chronos}~\cite{ansari2025chronos}: pretrained univariate, multivariate, and covariate-informed forecasting with group attention.
    \item \textbf{TimesFM}~\cite{das2024decoder}: a pretrained decoder-only Transformer using temporal patches for zero-shot forecasting.
\end{itemize}

\paragraph{LLM-based Methods.}
\begin{itemize}[leftmargin=*, noitemsep, topsep=2pt]
    \item \textbf{OFA}~\cite{zhou2023one}: adapts a pretrained language model by tuning selected layers while retaining frozen attention and feedforward layers.
    \item \textbf{Time-LLM}~\cite{jintime}: reprograms time series into language-model representations with prompt-based conditioning.
    \item \textbf{TimeReasoner}~\cite{cheng2025can}: uses LLM reasoning to assess temporal dynamics and produce forecasts.
    \item \textbf{Time-R1}~\cite{luo2025time}: combines supervised fine-tuning warmup with reinforcement learning for slow-thinking, multi-step temporal reasoning.
\end{itemize}

\paragraph{Agentic TSF Methods.}
\begin{itemize}[leftmargin=*, noitemsep, topsep=2pt]
    \item \textbf{AlphaCast}~\cite{zhang2025alphacast}: a training-free workflow combining temporal feature reasoning, knowledge and context, and reflection across multiple stages.
    \item \textbf{TimeSeriesScientist}~\cite{zhao2025timeseriesscientist}: coordinates specialized agents for diagnostics and preprocessing, model planning, fitting, validation and ensembling, and reporting.
\end{itemize}

\subsection{Baseline Implementation}
\label{app:baseline_implementation}
Deep learning baselines use their official implementations and Adam with a learning rate of $10^{-3}$ or $10^{-4}$ according to their recommended configurations. Early stopping uses patience 3. Inputs use Z-score standardization (zero mean and unit variance), and forecasts are returned to the raw scale for evaluation.

\FloatBarrier
\section{Agent Implementation}
\label{app:agent_implementation}
This section defines the diagnostic statistics, episode memory, and stage-aware prompt construction used by the forecasting harness.

\subsection{Diagnostic Tool Statistics}
\label{app:toolkit_statistics}
The toolkit in Section~\ref{subsec:toolkit} uses the following representative statistics. For a historical channel $v_{1:L}$, let $I$ denote its observed indices and \mbox{$n=|I|$}.

\paragraph{Data Quality Assessment.}
These tools identify missing values, saturation effects, and constant or abnormal channels. The missing-value ratio is $q_{\mathrm{miss}}=1-\frac{n}{L}$, helping distinguish temporal patterns from potential data issues.

\paragraph{Global Statistical Characterization.}
Global summaries describe central tendency, dispersion, correlation, and frequency-domain properties. Variability is summarized by the mean absolute deviation $D_{\mathrm{abs}}=\frac{1}{n}\sum_{i\in I}|v_i-\mu_v|$, where $\mu_v$ is the mean of observed values and $n>0$.

\paragraph{Structural and Dynamic Analysis.}
Within-series tools detect change points, stable segments, and local extrema. For a fully observed segment $[a,b]$, the average slope is $\operatorname{slope}_{[a,b]}=\frac{v_b-v_a}{b-a}$, where $a<b$, providing evidence about trends and regime shifts.

\paragraph{Event-level Summarization.}
Event tools summarize rising, declining, stable, and oscillatory behavior. Pattern prevalence is $\nu_c=\frac{n_c}{S}$, where $n_c$ counts segments labeled $c$ and $S>0$ is the total number of segments. The pattern with the largest prevalence summarizes the dominant behavior.

\paragraph{Model Adequacy and Residual Diagnostics.}
Residual tools assess how well a simple baseline explains history using $r_i=v_i-\breve v_i$, where $\breve v_i$ is a baseline fitted value on history. Residual distributions, autocorrelation, and tail behavior guide modeling and refinement decisions.

\subsection{Memory Update and State Assembly}
\label{app:memory_update}
For each episode, memory $M_k=(D_k,Q_k)$ stores \textbf{Analysis History} $D_k$: diagnostic summaries, structural indicators, and event descriptions; and \textbf{Prediction Results} $Q_k$: forecasts returned by model tools. Both components start empty and retain the current episode's tool results across turns. The environment manages this memory and constructs the next state as in Equation~\ref{eq:state_transition}. During RL training, multi-turn GRPO updates the shared decision policy. Historical observations and task context are supplied separately from these two memory components.

At each decision step, $P$ constructs a new policy-visible prompt $s_k=P(z,M_k)$ from task inputs and structured memory. The available diagnostics and forecasts determine the stage-specific instructions and permitted actions. Previous agent responses and raw tool messages are not additionally appended as conversation history; tool-derived evidence is carried forward through $D_k$ and $Q_k$. Early prompts contain the full historical window and task context. Refinement prompts combine recent observations, task context, and retained diagnostics and forecasts. EPF context contains forecast-horizon exogenous covariates and textual information available at the forecast origin; LTSF uses historical inputs.
\subsection{Stage-aware Prompts}
\label{app:stage_prompt}
\begin{itemize}[leftmargin=*, topsep=2pt]
    \item \textbf{Preparation:} when analysis is absent, supply the full historical window and task context and permit feature extraction and diagnostic tools. Independent tool calls may execute in parallel within one response.
    \item \textbf{Forecasting:} when analysis exists but no forecast is stored, supply historical inputs, task context, and Analysis History and permit forecasting model invocation with evidence-guided model selection.
    \item \textbf{Refinement and Output:} when diagnostics and forecasts are available, supply recent observations, task context, Analysis History, and Prediction Results and permit reasoning, refinement, and final output.
\end{itemize}
These labels identify logical stages rather than a fixed number of agent responses. Episodes follow the decision budget $K_{\max}$ defined in the main text. The environment constrains the stage order and permitted actions, while the policy selects tools, routes models, and revises predictions within this protocol.

\paragraph{Nominal Episode Procedure.}
The state transitions in Equation~\ref{eq:state_transition} can be summarized as follows:
\begin{enumerate}[leftmargin=*, noitemsep, topsep=2pt]
    \item Initialize $M_1=(\varnothing,\varnothing)$ for the supplied forecasting input $z$.
    \item At decision step $k\leq K_{\max}$, construct $s_k=P(z,M_k)$ and determine the permitted actions from the available diagnostics and forecasts.
    \item Generate $a_k\sim\pi_\theta(\cdot\mid s_k)$. If it provides the final forecast, terminate the episode with $\hat y(\tau)$.
    \item Otherwise execute its tool calls, obtain $o_k=T(s_k,a_k)$, update $M_{k+1}=U(M_k,a_k,o_k)$, and continue with the next state.
\end{enumerate}
During RL, the final forecast is scored against the training target $y$. At inference, the agent completes the procedure using the supplied inputs and tool outputs. This summary describes the nominal interaction; runtime budgets and failure handling are separate implementation settings.

\subsection{Prompt Template}
\label{sec:prompt_design}
Box~\ref{pro:agent_workflow} summarizes the stage protocol and representative tool calls. Inputs follow Appendices~\ref{app:task_configuration} and~\ref{app:input_access}. XML-style tags separate reflection from serialized timestamp--value predictions. This summary illustrates the workflow rather than reproducing a verbatim run prompt.

Negative wholesale electricity prices can be valid market outcomes~\cite{iea2025electricity}; the prompt therefore requires contextual evidence before revising them.

\definecolor{frameblue}{RGB}{25, 50, 120}
\definecolor{bgblue}{RGB}{235, 240, 255}

\newtcolorbox{StrategyBox}[3][frameblue]{
  enhanced,
  width=\linewidth,
  title={#3},
  colframe=#1,
  colback=#2,
  colbacktitle=#1,
  coltitle=white,
  fonttitle=\bfseries,
  fontupper=\rmfamily\small,
  arc=1.5mm,
  boxrule=1.2pt,
  top=2mm, bottom=2mm, left=2mm, right=2mm,
  toptitle=0.5mm, bottomtitle=0.5mm,
  before upper={\raggedright\setlength{\parindent}{0pt}}
}

\begin{nolinenumbers}
\refstepcounter{idx}
\label{pro:agent_workflow}
\begin{StrategyBox}[frameblue]{bgblue}{Box~\theidx: Forecasting Prompt Summary}

\noindent\textbf{Role \& Inputs:} Forecast the requested target and horizon from historical inputs. For EPF, also use forecast-horizon covariates and text available at the forecast origin.

\noindent\textbf{Preparation:} Assess data quality, distribution, dynamics, and events with diagnostic tools such as \texttt{extract\_basic\_statistics} and \texttt{extract\_within\_channel\_dynamics}. Store outputs in Analysis History. Use diagnostic tools only at this stage.

\noindent\textbf{Forecasting:} Use Analysis History to select ARIMA, Chronos, PatchTST, or iTransformer via \texttt{predict\_time\_series}. Store the forecast in Prediction Results.

\noindent\textbf{Refinement:} Use recent observations, task context, and both memory components to review and revise anomalous patterns. Negative electricity prices alone do not justify revision.

\noindent\textbf{Final Output:} Use \texttt{<think>...</think>} for reflection and \texttt{<answer>...</answer>} for timestamp--value predictions; return only these two blocks.

\end{StrategyBox}
\end{nolinenumbers}
\FloatBarrier

\section{Training Details}
\label{app:implementation}
\label{app:training_protocol}
We describe the computational setup, trajectory distillation, multi-turn RL, curriculum, and reward computation. Task and evaluation settings appear in Appendix~\ref{app:task_configuration}; baseline settings appear in Appendix~\ref{app:baseline_implementation}.

\subsection{Model and Computational Environment}
We train a single Qwen3-8B agent on samples from all ten datasets using PyTorch and Hugging Face Transformers. Both SFT and RL run on four NVIDIA A800 GPUs with 80GB memory per device. RL uses gradient checkpointing and BF16 precision for efficient large-scale training. Table~\ref{tab:hyperparameters} summarizes SFT and GRPO settings.

\begin{table}[!htbp]
    \centering
    \caption{Hyperparameter settings for Cast-R1 training.}
    \label{tab:hyperparameters}
    \footnotesize
    \setlength{\tabcolsep}{3pt}
    \begin{tabular*}{\linewidth}{@{\extracolsep{\fill}}l|cc}
    \toprule
    \textbf{Hyperparameter} & \textbf{SFT} & \textbf{GRPO} \\
    \midrule
    Base Model & \multicolumn{2}{c}{Qwen3-8B} \\
    Precision & \multicolumn{2}{c}{BF16} \\
    Optimizer & AdamW & AdamW \\
    Learning Rate & $1.0 \times 10^{-5}$ & $1.0 \times 10^{-6}$ \\
    LR Scheduler & Cosine & Cosine \\
    Epochs & 1 & 1 \\
    Global Batch Size & 16 & 64 \\
    Group Size ($G$) & - & 8 \\
    Max Prompt Length & 4096 & 8192 \\
    Max Response Length & 2048 & 4096 \\
    Temperature & - & 1.0 \\
    KL Coefficient ($\beta$) & - & 0.04 \\
    Gradient Accumulation & 1 & 4 \\
    \bottomrule
    \end{tabular*}
\end{table}

\subsection{Trajectory Distillation}
\label{app:sft_protocol}
\paragraph{Teacher Demonstrations.} A DeepSeek-V3.2 teacher generates multi-turn trajectories using the forecasting agent harness, with forecasting windows drawn exclusively from the training splits of the ten datasets. The trajectories cover diagnostic tool invocation, forecasting model selection, and prediction refinement.

\subsubsection{Trajectory Evaluation and Selection}
\label{app:trajectory_selection}
We apply a two-stage filtering procedure to the teacher-generated trajectories.

\paragraph{Execution Validity.} We first check whether each trajectory completes the required tool interactions and produces a valid forecast. Trajectories with failed tool execution, malformed outputs, or non-finite prediction values are discarded.

\paragraph{Forecasting Quality.} For each valid trajectory $\tau$, we evaluate its final prediction against the ground-truth target from the training split. For forecasting horizon $H$, the error is:
\begin{equation}
E(\tau)=\frac{1}{H}\sum_{h=1}^{H}
\bigl(\hat{y}_{h}(\tau)-y_h\bigr)^2.
\label{eq:trajectory_selection_mse}
\end{equation}
Within each dataset, valid trajectories are ranked by ascending forecasting MSE. We retain the 120 lowest-error complete trajectories per dataset, yielding 1,200 demonstrations with equal dataset representation in $D_{\mathrm{SFT}}$.

Trajectory selection evaluates completed teacher interactions. The curriculum instead scores forecasting inputs to determine sample difficulty and training order (Appendix~\ref{app:curriculum}).

\subsubsection{Supervised Distillation}
Each retained demonstration pairs a teacher response with its decision state; accumulated tool feedback supplies context for subsequent responses. The supervised targets in Equation~\ref{eq:sft_objective} are the teacher responses containing tool calls or the final forecast. The Qwen3-8B student is fine-tuned for one epoch with AdamW before RL.
For a retained trajectory with $K(\tau^\star)$ responses, the corresponding supervision contains $K(\tau^\star)$ state--response pairs. Thus, the response-example count is:
\begin{equation}
N_{\mathrm{resp}}=\sum_{\tau^\star\in D_{\mathrm{SFT}}}K(\tau^\star).
\label{eq:response_count}
\end{equation}
This distinguishes the number of complete demonstrations from the number of supervised responses. In autoregressive form,
\begin{equation}
\begin{aligned}
&\log\pi_\theta(a_k^\star\mid s_k^\star)\\
&\quad=\sum_{j=1}^{|a_k^\star|}\log\pi_\theta(a_{k,j}^\star\mid s_k^\star,a_{k,<j}^\star).
\end{aligned}
\label{eq:response_supervision}
\end{equation}
The state contains the task inputs and available tool observations, while the response tokens supply the likelihood terms in Equation~\ref{eq:sft_objective}.

\subsection{Multi-turn Reinforcement Learning}
\label{app:rl_protocol}
Multi-turn GRPO samples $G=8$ trajectories per input and computes group-relative advantages from terminal rewards. Sampling uses temperature 1.0. The shared training configuration is listed in Table~\ref{tab:hyperparameters}.
For trajectory $i$, response $k$, and response token $j$, let $h_{i,k,j}=(s_{i,k},a_{i,k,<j})$. The token-level expansion of the outcome-supervised GRPO formulation~\cite{shao2024deepseekmath} uses:
\begin{equation}
\rho_{i,k,j}(\theta)=\frac{\pi_\theta(a_{i,k,j}\mid h_{i,k,j})}{\pi_{\mathrm{old}}(a_{i,k,j}\mid h_{i,k,j})}
\label{eq:token_policy_ratio}
\end{equation}
and the clipped surrogate:
\begin{equation}
\begin{aligned}
\ell_{i,k,j}(\theta)=\min\bigl\{&\rho_{i,k,j}(\theta)\hat A_i,\\
&\operatorname{clip}(\rho_{i,k,j}(\theta),1-\epsilon,1+\epsilon)\hat A_i\bigr\},
\end{aligned}
\label{eq:clipped_response_surrogate}
\end{equation}
where $\epsilon$ is the clipping parameter and $\hat A_i$ is the trajectory advantage in Equation~\ref{eq:group_advantage}. The same outcome advantage applies across the trajectory's agent responses, with KL regularization against a reference policy. Tool outputs condition the next state as environment observations. Rewards within a group are compared for the same input and target; they are not pooled across unrelated tasks.
\begin{table*}[t]
	\centering
	\footnotesize
	\setlength{\tabcolsep}{1pt}
	\newcommand{\maeresultwidth}{\dimexpr(\linewidth-82pt-22\tabcolsep-2\arrayrulewidth)/10\relax}
	\begin{tabular}{>{\centering\arraybackslash}m{82pt}|*{5}{>{\centering\arraybackslash}m{\maeresultwidth}}|*{5}{>{\centering\arraybackslash}m{\maeresultwidth}}}
		\toprule
		& \multicolumn{5}{c|}{\shortstack{\textbf{LTSF (Historical Inputs)}\\($L=96$, $H=96$)}} & \multicolumn{5}{c}{\shortstack{\textbf{EPF (Forecast-Origin Context)}\\($L=168$, $H=24$)}} \\
		\cmidrule(lr){2-6}\cmidrule(lr){7-11}
		Model & \multicolumn{1}{c}{ETTh1} & \multicolumn{1}{c}{ETTh2} & \multicolumn{1}{c}{ETTm1} & \multicolumn{1}{c}{ETTm2} & \multicolumn{1}{c|}{Wind} & \multicolumn{1}{c}{BE} & \multicolumn{1}{c}{DE} & \multicolumn{1}{c}{FR} & \multicolumn{1}{c}{NP} & \multicolumn{1}{c}{PJM} \\
		\midrule
		ARIMA & 2.429 & 3.248 & 2.513 & 4.325 & 28.767 & 13.549 & 11.150 & 13.224 & 4.116 & 8.481 \\
		Prophet & 4.592 & 3.548 & 2.504 & 3.781 & 56.059 & 14.988 & 12.238 & 14.422 & 3.941 & 5.739 \\
		PatchTST & 1.638 & 2.008 & 1.397 & 1.945 & 20.465 & \underline{10.400} & \underline{8.029} & \underline{8.287} & 3.474 & 4.103 \\
		iTransformer & 1.513 & \underline{1.930} & 1.460 & \underline{1.661} & 18.186 & 14.375 & 11.230 & 12.618 & 4.039 & 5.501 \\
		ConvTimeNet & \underline{1.439} & 2.578 & \underline{1.252} & 2.057 & 18.343 & 10.804 & 9.405 & 8.993 & 3.365 & 4.313 \\
		TimeXer & 2.204 & 2.663 & 1.609 & 3.113 & 31.914 & 12.286 & 10.274 & 9.483 & 3.490 & 4.252 \\
		DLinear & 2.239 & 2.558 & 1.751 & 2.509 & 29.325 & 12.590 & 10.619 & 9.735 & 3.896 & 4.508 \\
		Chronos & 2.250 & 2.820 & 1.323 & 2.117 & 17.182 & 11.129 & 8.584 & 8.423 & \textbf{3.155} & \underline{3.712} \\
		TimesFM & 2.223 & 2.231 & 2.012 & 2.114 & 20.898 & 11.284 & 10.032 & 9.932 & 4.667 & 5.001 \\
		OFA & 2.917 & 3.895 & 2.771 & 3.152 & 26.557 & 13.665 & 14.235 & 13.505 & 4.462 & 4.913 \\
		Time-LLM & 2.500 & 2.979 & 1.668 & 3.352 & 18.775 & 10.997 & 9.927 & 10.975 & 5.221 & 4.139 \\
		TimeReasoner & 1.945 & 2.551 & 2.552 & 2.925 & 22.559 & 11.976 & 15.223 & 15.997 & 6.661 & 4.312 \\
		Time-R1 & 2.136 & 2.599 & 1.376 & 3.351 & \underline{15.110} & 10.710 & 13.614 & 14.307 & 5.957 & 3.856 \\
		\mbox{TimeSeriesScientist} & 3.022 & 3.972 & 1.921 & 3.949 & 25.648 & 14.939 & 10.388 & 12.71 & 4.156 & 5.388  \\
		AlphaCast & 2.345 & 2.924 & 1.733 & 3.566 & \textbf{13.065} & 10.713 & 10.228 & 9.103 & \underline{3.191} & \textbf{3.586}  \\
		\midrule
		\rowcolor[gray]{0.95} \textbf{Cast-R1} & \textbf{1.320} & \textbf{1.747} & \textbf{1.160} & \textbf{1.581} & 16.125 & \textbf{10.332} & \textbf{7.400} & \textbf{8.105} & 3.255 & 3.877 \\
		\bottomrule
	\end{tabular}
		\caption{MAE results for the same benchmark families and input regimes as Table~\ref{tab:main_results}. The groups do not constitute a controlled context ablation; input access follows Appendix~\ref{app:input_access}. Lower values are better. Best values are \textbf{bold}; second-best values are \underline{underlined}.}
	\label{tab:main_results_mae}
\end{table*}

\subsection{Curriculum Learning}
\label{app:curriculum}
The curriculum combines the prediction MSE of a reference forecaster with normalized permutation entropy~\cite{bandt2002permutation} of the historical series. Prediction error measures difficulty relative to that forecaster, while permutation entropy characterizes the series' intrinsic complexity. These signals guide filtering of noise-dominated samples and the progressive introduction of retained samples from simpler trends and seasonal patterns to more complex dynamics. The reference forecaster used for difficulty scoring has a separate role from the DeepSeek-V3.2 teacher that generates interaction trajectories. The trajectory selection procedure in Appendix~\ref{app:trajectory_selection} determines which completed teacher interactions provide SFT supervision, whereas the curriculum operates on forecasting inputs to determine sample difficulty and training order.

For a scalar sequence $v_{1:L}$, embedding order $m\geq2$, and delay $\ell\geq1$ with $L>(m-1)\ell$, ordinal patterns are obtained from vectors $(v_t,v_{t+\ell},\ldots,v_{t+(m-1)\ell})$. Let $p_\sigma$ be the frequency of pattern $\sigma$ among the $m!$ possible orderings in $\Pi_m$. Normalized permutation entropy is:
\begin{equation}
\mathrm{PE}_{m,\ell}(v)=-\frac{1}{\log(m!)}\sum_{\sigma\in\Pi_m}p_\sigma\log p_\sigma,
\label{eq:normalized_permutation_entropy}
\end{equation}
with $0\log0=0$ and $\mathrm{PE}_{m,\ell}\in[0,1]$. Higher values indicate a more uniform distribution of local ordinal patterns. This statistic measures a different property from prediction error, which depends on the reference forecaster.

\subsection{Reward Implementation}
\label{app:reward_details}
\paragraph{Output Format.} Both \texttt{<answer>} and \texttt{</answer>} are required. If either tag is missing, reward computation stops and returns $-1$. The tags act as a validity gate; outputs passing this check are scored by Equation~\ref{eq:forecast_reward}.

\paragraph{Accuracy and Temporal Structure.} Prediction MSE uses predictions and targets normalized by the same target mean and standard deviation. This target-dependent normalization is used only in training reward computation; the reported MSE and MAE use the raw scale. Each nonnegative prediction, seasonal, or trend error $e$ is mapped to $q(e)=[1+\log(1+e)]^{-1}$. Moving averages decompose both sequences into trend and seasonal components. Failed decomposition contributes zero for the affected component. Local maxima and minima are detected with a sliding window and matched within $\pm2$ time steps; their hit rates supply the alignment scores.
For aligned $H$-step sequences and a positive target standard deviation $\sigma_y$, the common normalization gives
\begin{equation}
e_{\mathrm{pred}}=\frac{1}{H\sigma_y^2}\sum_{i=1}^{H}(\hat y_i-y_i)^2.
\label{eq:normalized_prediction_error}
\end{equation}
Consequently, a target and its forecast are compared on a common scale; they are not standardized independently. The score mapping has $q(0)=1$ and decreases with error. The training reward and the raw-scale evaluation metrics therefore serve distinct roles.

\paragraph{Weights and Completeness.} Table~\ref{tab:reward_components} lists the terms in Equation~\ref{eq:forecast_reward}. The length factor $p=\min(1,n_{\mathrm{pred}}/n_{\mathrm{gt}})$ scales the entire weighted sum, including the constant $w_{\mathrm{len}}$. It penalizes outputs shorter than the target and saturates at one for longer outputs; by itself, it does not enforce exactly $H$ predictions. For component scores $u_j\in[0,1]$, the weights give $0\leq R_{\mathrm{valid}}\leq1.2p\leq1.2$.

\begin{table}[!htbp]
    \centering
    \caption{Components of the valid-output reward before multiplication by the length factor $p$.}
    \label{tab:reward_components}
    \footnotesize
    \setlength{\tabcolsep}{3pt}
    \begin{tabular*}{\linewidth}{@{\extracolsep{\fill}}llr}
    \toprule
    \textbf{Component} & \textbf{Score} & \textbf{Weight} \\
    \midrule
    Length constant & $1$ & $0.10$ \\
    Prediction accuracy & $q(e_{\mathrm{pred}})$ & $0.70$ \\
    Seasonality & $q(e_{\mathrm{seas}})$ & $0.05$ \\
    Trend & $q(e_{\mathrm{trend}})$ & $0.15$ \\
    Maximum alignment & Maximum hit rate & $0.10$ \\
    Minimum alignment & Minimum hit rate & $0.10$ \\
    \bottomrule
    \end{tabular*}
\end{table}

\FloatBarrier
\section{Additional Experimental Results}
\label{app:detailed_results}
We first report the full MAE comparison, then examine diagnostic tools, forecasting models, planning, and refinement. A qualitative forecast comparison complements these aggregate results.

\subsection{MAE Results and Standalone Predictor Comparison}
\label{app:main_mae}
Table~\ref{tab:main_results_mae} complements the MSE comparison in Table~\ref{tab:main_results}. \model{} achieves the lowest MAE on seven datasets. AlphaCast achieves the lowest MAE on Wind and PJM, while Chronos leads on NP. On Wind, \model{} ranks first in MSE and third in MAE, showing that method rankings differ across the two error criteria.
Comparing \model{} with the best individual baseline among the four candidate forecasters (ARIMA, Chronos, PatchTST, and iTransformer) gives lower MSE on all ten datasets. The per-dataset reductions range from 0.95\% to 21.50\%, averaging 12.09\%. These numbers describe the complete agent-and-tool pipeline relative to standalone predictors; they do not isolate the contribution of model routing from refinement.

\subsection{Diagnostic Tool Ablations}
\label{app:toolkit_analysis}
Table~\ref{tab:feature_tools} compares diagnostic removals on ETTh1 and NP. Each variant removes the corresponding diagnostic group defined in Section~\ref{subsec:toolkit} and Appendix~\ref{app:toolkit_statistics}. Removing basic statistics causes the largest reported MSE increase on both datasets. All diagnostic removals worsen ETTh1 results. On NP, removing data quality assessment reduces MSE by 4.82\% and MAE by 0.86\%, while removing residual diagnostics or channel dynamics improves MSE only. These comparisons show dataset- and metric-dependent contributions of individual diagnostic groups.

\begin{table}[!htbp]
    \centering
    \caption{Diagnostic tool ablations on ETTh1 and NP.}
    \footnotesize
    \renewcommand{\arraystretch}{1.15}
    \setlength{\tabcolsep}{2.5pt}
    \begin{tabular*}{\linewidth}{@{\extracolsep{\fill}}lrrrr}
        \toprule
        \multirow{2}{*}{\textbf{Variant}} & \multicolumn{2}{c}{\textbf{ETTh1}} & \multicolumn{2}{c}{\textbf{NP}} \\
         & \textbf{MSE} & \textbf{MAE} & \textbf{MSE} & \textbf{MAE} \\ \midrule
        w/o Basic Stats & 9.814 & 2.384 & 36.781 & 4.715 \\
        w/o Channel Dynamics & 7.012 & 1.532 & 24.711 & 3.386 \\
        w/o Residual Diagnostics & 6.543 & 1.341 & 23.774 & 3.335 \\
        w/o Data Quality & 6.942 & 1.574 & \textbf{23.557} & \textbf{3.227} \\
        w/o Event Patterns & 7.571 & 1.642 & 27.154 & 3.646 \\ \midrule
        \textbf{Cast-R1} & \textbf{6.062} & \textbf{1.320} & 24.750 & 3.255 \\
        \bottomrule
    \end{tabular*}
    \label{tab:feature_tools}
\end{table}
\begin{figure*}[!t]
    \centering
    \includegraphics[width=\linewidth]{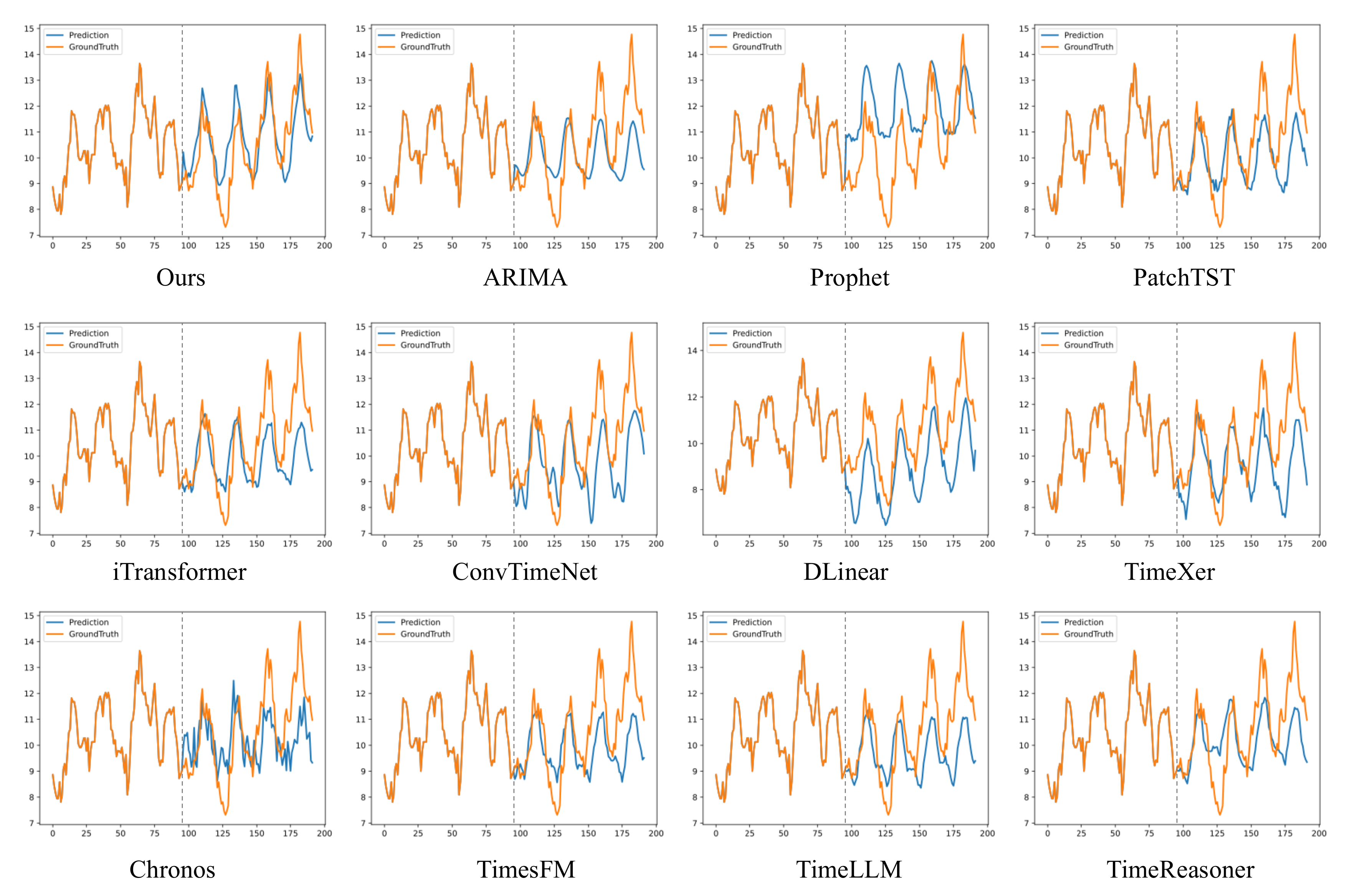}
    \caption{Forecast comparison on ETTh1 oil temperature with $L=96$ and $H=96$. Dashed lines separate the 96 historical observations from the 96-step forecasting horizon.}
    \label{fig:vis_comparison}
\end{figure*}
\subsection{Forecasting Model Ablations}
Table~\ref{tab:model_tools} compares individual forecasting model removals on ETTh1 and NP. Each variant removes the named forecaster from the candidate pool $B$ in Section~\ref{subsec:toolkit}. Every tested removal increases MSE and MAE on both datasets. Removing Chronos produces the largest reported MSE increase on NP, whereas removing PatchTST has the strongest effect on ETTh1. These results show dataset-dependent sensitivity to the available forecasters and favor retaining the full candidate pool in the two evaluated settings.

\begin{table}[!htbp]
    \centering
    \caption{Forecasting model ablations on ETTh1 and NP.}
    \footnotesize
    \renewcommand{\arraystretch}{1.15}
    \setlength{\tabcolsep}{2.5pt}
    \begin{tabular*}{\linewidth}{@{\extracolsep{\fill}}lrrrr}
        \toprule
        \multirow{2}{*}{\textbf{Variant}} & \multicolumn{2}{c}{\textbf{ETTh1}} & \multicolumn{2}{c}{\textbf{NP}} \\
         & \textbf{MSE} & \textbf{MAE} & \textbf{MSE} & \textbf{MAE} \\ \midrule
        w/o ARIMA & 6.614 & 1.413 & 24.788 & 3.334 \\
        w/o Chronos & 8.414 & 1.622 & 55.418 & 7.097 \\
        w/o PatchTST & 14.261 & 3.014 & 37.191 & 4.277 \\
        w/o iTransformer & 13.512 & 2.825 & 30.387 & 4.054 \\ \midrule
        \textbf{Cast-R1} & \textbf{6.062} & \textbf{1.320} & \textbf{24.750} & \textbf{3.255} \\
        \bottomrule
    \end{tabular*}
    \label{tab:model_tools}
\end{table}

\subsection{Planning and Refinement Ablations}
Table~\ref{tab:agent_components} compares planning and refinement removals on ETTh1 and NP. Planning refers to diagnostic tool selection and evidence-guided model routing, while refinement refers to numerical revision of the intermediate forecast, following the workflow in Section~\ref{sec:method}. These are component ablations within the prescribed stage order in Appendix~\ref{app:stage_prompt}. Removing either component increases MSE and MAE, with planning removal causing the larger deterioration on both datasets. Relative to the variant without refinement, the full framework reduces MSE by 33.69\% on ETTh1 and 2.61\% on NP. The reported gains therefore differ in magnitude across these two datasets.

\begin{table}[!htbp]
    \centering
    \caption{Agent component ablations on ETTh1 and NP, removing planning or refinement.}
    \footnotesize
    \renewcommand{\arraystretch}{1.15}
    \setlength{\tabcolsep}{2.5pt}
    \begin{tabular*}{\linewidth}{@{\extracolsep{\fill}}lrrrr}
        \toprule
        \multirow{2}{*}{\textbf{Variant}} & \multicolumn{2}{c}{\textbf{ETTh1}} & \multicolumn{2}{c}{\textbf{NP}} \\
         & \textbf{MSE} & \textbf{MAE} & \textbf{MSE} & \textbf{MAE} \\ \midrule
        w/o Planning & 18.184 & 3.914 & 38.540 & 5.549 \\
        w/o Refinement & 9.142 & 1.542 & 25.413 & 3.841 \\ \midrule
        \textbf{Cast-R1} & \textbf{6.062} & \textbf{1.320} & \textbf{24.750} & \textbf{3.255} \\
        \bottomrule
    \end{tabular*}
    \label{tab:agent_components}
\end{table}

\subsection{Forecast Visualization}
Figure~\ref{fig:vis_comparison} compares \model{} with selected baseline forecasters on an ETTh1 oil-temperature segment with $L=96$ and $H=96$, recurring oscillations, and a sharp downward excursion. \model{} retains the oscillatory pattern and tracks several later peaks, although it does not fully capture the deepest drop and some peak amplitudes. Other forecasters differ in timing, level, and amplitude, with several curves smoothing peaks or shifting troughs. This example illustrates local tracking behavior and remaining extreme-event errors; aggregate performance is reported in Tables~\ref{tab:main_results} and~\ref{tab:main_results_mae}.

\end{document}